\documentclass[journal]{IEEEtran}
\IEEEoverridecommandlockouts

\usepackage{cite}
\usepackage{amsmath,amssymb,amsfonts}
\usepackage{algorithmic}
\usepackage{graphicx}
\usepackage{textcomp}
\usepackage{xcolor}
\usepackage{booktabs}
\usepackage{multirow}
\usepackage{array}
\usepackage[colorlinks,linkcolor=red,anchorcolor=red,citecolor=blue]{hyperref}
\usepackage[lined,boxed,ruled]{algorithm2e}

\graphicspath{{./figs/}}

\def\BibTeX{{\rm B\kern-.05em{\sc i\kern-.025em b}\kern-.08em
    T\kern-.1667em\lower.7ex\hbox{E}\kern-.125emX}}

\begin{document}

\title{A Large-Scale Empirical Evaluation of MMAO Under Fair-Budget Continuous and Discrete Benchmarks}

\author{Jinliang Xu$^{*}$ and Liping Ma%
\thanks{Jinliang Xu received the bachelor's and master's degrees from Beijing University of Posts and Telecommunications, Beijing, China, where he also received the Ph.D. degree in computer science in 2020. He is with the China Academy of Information and Communications Technology, Beijing, China. His current research interests include heuristic optimization, agent-based applications, and blockchain systems. His e-mail: jlxufly@gmail.com and xujinliang@caict.ac.cn.}%
\thanks{Liping Ma is with the Department of Disease Control and Prevention, The Seventh Medical Center of Chinese PLA General Hospital, Beijing, China; e-mail: lipingmaqzx@163.com.}}

\maketitle

\begin{abstract}
This paper evaluates the Metabolic Multi-Agent Optimizer (MMAO) under a stricter empirical protocol rather than reintroducing the framework itself. The study asks whether MMAO's closed-loop resource-allocation principle remains credible under broader, more standard, and more explicitly budget-controlled continuous and discrete benchmarks. The main completed matrix covers eight CEC2017 functions at 10D and 30D with 20 seeds each, and five TSPLIB instances with 20 seeds each, together with stronger reproducible baselines including PSO-lite, ES-lite, and an iterated-greedy 2-opt route baseline. We further add trajectory-level diagnostics for communal budget, success rate, role evolution, and population turnover, plus an auxiliary OR-Library multiple-knapsack slice to extend the discrete evidence beyond routing. Under this protocol, MMAO clearly outperforms the external baseline set on the continuous side and on the TSPLIB side, while the ablation variants remain much closer to the full method than the external baselines are. The results support endogenous resource redistribution across the tested domains, while sharper mechanism isolation and broader competition-grade comparison remain open.
\end{abstract}


\section{Introduction}
\IEEEPARstart{M}{etaheuristic} research often advances through two distinct but easily conflated contributions: proposing a new adaptive search principle and demonstrating that the principle survives strong evidence pressure. Many papers do the first more successfully than the second. Mechanisms are introduced, benchmark slices are selected, and preliminary results are obtained, yet the supporting experimental protocol remains too narrow to establish how robust the claimed contribution really is across problem classes, dimensions, and random seeds \cite{stripinis2024benchmarking, vermetten2024large, cenikj2025comparing}.

The Metabolic Multi-Agent Optimizer (MMAO) was introduced as a framework in which search intensity, role evolution, population turnover, and redistribution of effort are financed through a private-public metabolic loop rather than through externally attached schedules \cite{xu2026mmao}. A contraction study then asked a different question: what is the smallest endogenous closed-loop structure that still meaningfully counts as MMAO \cite{xu2026minimalmmao}. Those studies clarified the conceptual identity of the method, but they did not yet settle the central empirical question facing any new optimization framework: does the closed-loop resource-allocation principle remain competitive under larger and more standard evaluation protocols. This question sits within a much older metaheuristic tradition ranging from classical evolutionary adaptation and simulated annealing to more recent metaphor-driven optimizers \cite{holland1975adaptation, kirkpatrick1983optimization, li2020slime, kiran2015tsa}.

This paper addresses that question directly. We test whether MMAO remains competitive as benchmark coverage, budget normalization, repetition policy, and statistical reporting are tightened. The emphasis is therefore on \emph{evaluation design}: broader benchmark coverage, explicit budget control, repeated trials, aggregate statistics, and mechanism-aware ablations.

Our working hypothesis is conservative. If the MMAO idea has genuine value, it should appear not as isolated wins on favorable tasks, but as a repeatable pattern under matched budgets across both continuous and discrete search spaces. If the framework instead relies mainly on hidden tuning or narrow engineering choices, that advantage should erode once the protocol becomes broader and stricter.

The contributions of this paper are experimental and methodological:
\begin{enumerate}
    \item We define and execute a fair-budget evaluation protocol for MMAO across continuous and discrete optimization, including benchmark coverage, repetition policy, stronger reproducible baselines, and rank-based statistical reporting.
    \item We specify the MMAO configuration used in this study, separating the experimentally evaluated algorithmic realization from the broader conceptual description developed in prior work.
    \item We design a mechanism-centered ablation suite that tests whether performance differences can be attributed to communal resource sharing, continuous role adaptation, success feedback, and lifecycle turnover.
    \item We report benchmark results and open mechanism-level questions, and complement the score tables with trajectory-level diagnostics of the metabolic loop.
\end{enumerate}

The remainder of the paper is organized as follows. Section II reviews evaluation-oriented literature relevant to adaptive metaheuristics and fair benchmarking. Section III describes the MMAO configuration used for strong validation. Section IV presents the experimental protocol, including benchmark suites, budgets, baselines, and statistical procedures. Section V reports the completed continuous and discrete benchmark results. Section VI analyzes ablations and mechanism diagnostics. Section VII discusses interpretation boundaries and threats to validity. Section VIII concludes.

\section{Related Work}
The present study is situated less in the line of ``new metaphor'' algorithms than in the line of \emph{evaluation discipline} for adaptive heuristics. Three strands of prior work are particularly relevant.

First, the parameter-control literature has long emphasized the difference between exogenous schedules and endogenous adaptation \cite{eiben1999parameter, karafotias2015parameter}. Classical evolutionary and swarm methods such as PSO, DE, CMA-ES, and their descendants differ not only in operator design, but also in how adaptation is internalized \cite{kennedy1995particle, Hansen2001, Tanabe2014, Stanovov2018, brest2006self, omeradzic2024self}. Derivative-free continuous search also provides a complementary perspective in which adaptation is reflected through sampling scale and gradient-free probing rather than only through population operators \cite{nesterov2017random}. This is directly relevant to MMAO because its main claim is not a specific move operator, but a shared control economy for search effort.

Second, modern benchmark studies increasingly stress that isolated function-level wins are a weak basis for methodological conclusions. Large benchmark comparisons, repeated-seed reporting, performance profiles, and nonparametric aggregate statistics have become important safeguards against overinterpretation \cite{stripinis2024benchmarking, vermetten2024large, cenikj2025comparing, raponi2023optimizing, antipov2024already}. In continuous optimization, such standards are often operationalized through CEC-style suites with multiple dimensions and substantial function-evaluation budgets. In combinatorial optimization, a parallel discipline exists around established libraries such as TSPLIB and around comparison to classical high-quality heuristics \cite{Helsgaun2017, Voudouris1999, ye2023deepaco, elorza2024transforming}. The multiobjective benchmarking literature also reinforces the need to distinguish true algorithmic advantage from artefacts of ranking and diversity handling \cite{doerr2023understanding, zheng2024approximation}.

Third, adaptive population sizing and online effort redistribution have become recurring themes across optimization research \cite{doerr2025speeding, li2022distributed, liu2022cooperative, cho2025configx, zhang2025laos, dong2025effective, li2014adaptivebandit, jiang2023knowledge}. Related work on dynamic and large-scale optimization also shows that transfer, subspace restriction, and perturbation scheduling can materially affect how limited search resources are used over time \cite{liu2023transfer, signorelli2025perturbation}. What is less common is a formulation in which population turnover, local intensity, and redistributive behavior are all generated by one bounded resource loop rather than by partially independent adaptation modules. MMAO belongs to this narrower family of endogenous resource-allocation approaches.

This study tests whether a cross-domain closed-loop allocation principle remains competitive under a broader, budget-controlled evaluation protocol.

\section{MMAO Configuration for Large-Scale Evaluation}
This paper evaluates a mature MMAO realization derived from the original cross-domain formulation \cite{xu2026mmao} rather than the deliberately contracted version studied in \cite{xu2026minimalmmao}. The intent is to test a configuration that is rich enough to represent the framework fairly, yet structured enough that its evaluation remains interpretable.

\subsection{Core State and Control Variables}
At iteration $t$, MMAO maintains an active population $\mathcal{P}_t=\{A_1,\ldots,A_{N_t}\}$ and a communal state. Each agent carries
\begin{equation}
    A_i(t)=\big(x_i(t), E_i(t), \phi_i(t), m_i(t), \tau_i(t)\big),
\end{equation}
where $x_i(t)$ is the current candidate solution, $E_i(t)$ is bounded private energy, $\phi_i(t)\in[0,1]$ is a continuous role state, $m_i(t)$ is local memory, and $\tau_i(t)$ tracks local age or stagnation. The population maintains a communal budget $B_t$, a recent progress scale $s_t$, and a recent success statistic $\rho_t$.

The defining design constraint is preserved from earlier MMAO work: adaptive behavior must be derived from the metabolic resource loop rather than attached as an external controller. In practice, this means that search scale, sensing intensity, branching tendency, pruning pressure, and reinvestment are all functions of the same normalized reward-and-cost accounting pathway.

\subsection{Continuous Instantiation}
In continuous domains, the evaluated MMAO variant uses symmetric zero-order probing, role-conditioned motion, bounded uphill tolerance, and budget-driven offspring placement around productive regions. The role state interpolates between broader exploratory probing and denser local refinement. The communal pool affects whether local gains are reinvested into offspring or retained as private survival capital.

\subsection{Discrete Instantiation}
In permutation domains, the same control loop is realized through route-level local improvement, structural perturbation, energy-aware acceptance, and lightweight shared edge memory. Rather than importing a domain-specific master heuristic, the implementation keeps the control logic aligned with the continuous side: resource gains fuel selective reinforcement, while stagnation and depletion trigger redistribution or replacement.

\subsection{Why This Configuration}
The large-scale study must evaluate a version of MMAO that is representative without becoming opaque. For that reason, the present configuration retains:
\begin{itemize}
    \item bounded private energy and a communal pool;
    \item normalized reward and recent-success feedback;
    \item continuous role adaptation rather than fixed species;
    \item dynamic population turnover through branching, pruning, and respawn;
    \item domain-aware continuous and discrete sensing operators.
\end{itemize}
It does not claim that this is the final or uniquely correct MMAO realization. It is simply the version judged most suitable for strong empirical validation at the current stage of the research line.

\begin{figure}[t]
\centering
\includegraphics[width=\columnwidth]{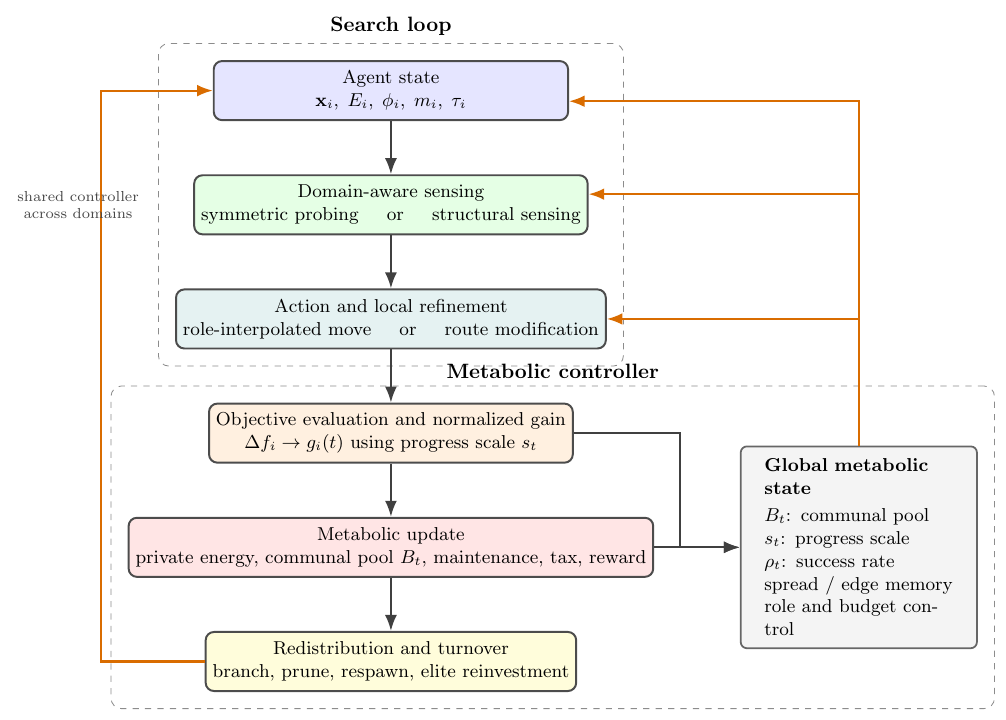}
\caption{MMAO as evaluated in the present study. Agent behavior, resource accounting, and lifecycle turnover remain coupled through a shared metabolic loop.}
\label{fig:architecture}
\end{figure}

\section{Experimental Protocol}
The protocol specifies benchmark coverage, budgets, and reporting procedures to support reproducible evaluation (see GitHub repository \texttt{mmao}\footnote{\url{https://github.com/wolfbrother/mmao}} or PyPI package \texttt{mmao-opt}\footnote{\url{https://pypi.org/project/mmao-opt/}}).

\subsection{Research Questions}
The study is organized around five questions:
\begin{enumerate}
    \item \textbf{RQ1:} Does MMAO remain competitive on standard continuous benchmarks across multiple function classes and dimensions?
    \item \textbf{RQ2:} Does MMAO remain competitive on standard discrete benchmarks across multiple TSPLIB instances and scales?
    \item \textbf{RQ3:} Are MMAO results stable across seeds under fixed budget conditions?
    \item \textbf{RQ4:} Can observed performance be attributed primarily to the closed-loop resource-allocation mechanism rather than to isolated implementation details?
    \item \textbf{RQ5:} What failure modes or boundary conditions become visible under stronger benchmark pressure?
\end{enumerate}

\subsection{Continuous Benchmark Suite}
The continuous side is designed around a CEC-style subset that spans unimodal, multimodal, hybrid, and composition functions. The completed strong-validation matrix is:
\begin{itemize}
    \item dimensions: $10D$ and $30D$;
    \item eight benchmark functions in the present completed study;
    \item 20 seeds per function-dimensionality pair;
    \item one fixed MMAO configuration family and one fixed budget rule per external baseline family within each dimension.
\end{itemize}

The present subset contains \texttt{F12017}, \texttt{F32017}, \texttt{F42017}, \texttt{F52017}, \texttt{F62017}, \texttt{F72017}, \texttt{F92017}, and \texttt{F102017}. This selection preserves a practical spread across relatively simple, multimodal, and hybrid-like difficulty while keeping the completed matrix computationally manageable.

\subsection{Discrete Benchmark Suite}
The discrete side focuses on TSPLIB because it provides widely recognized instances with reliable reference optima. The completed study uses the following five instances:
\begin{itemize}
    \item small-to-medium instances: \texttt{eil51}, \texttt{berlin52}, \texttt{st70};
    \item medium instances: \texttt{eil76}, \texttt{kroA100}.
\end{itemize}
Each instance is run with 20 seeds and evaluated under explicit route-construction and bounded local-search effort rather than wall-clock time, so that replication is less sensitive to hardware and implementation differences.

To avoid limiting the discrete evidence to routing only, we also include an auxiliary OR-Library 0/1 multiple-knapsack slice. This auxiliary slice is smaller than the TSPLIB matrix and is used conservatively as a cross-family check rather than as a co-equal primary benchmark campaign. Its role is to test whether the same closed-loop controller can still produce sensible behavior on binary combinatorial structure beyond tours.

\subsection{Baselines}
Baselines were selected to form a reproducible but materially stronger comparison layer that could still be executed across the full matrix within the current project scope.

For continuous optimization, the implemented baselines are:
\begin{itemize}
    \item \texttt{RandomSearch};
    \item \texttt{HillClimb};
    \item \texttt{DE-lite};
    \item \texttt{PSO-lite};
    \item \texttt{ES-lite}.
\end{itemize}

For discrete optimization, the implemented baselines are:
\begin{itemize}
    \item \texttt{NN+2opt};
    \item \texttt{RR-2opt};
    \item \texttt{CI+2opt};
    \item \texttt{IG-2opt}.
\end{itemize}

For the auxiliary multiple-knapsack slice, we use \texttt{GreedyDensity}, \texttt{RandomRepair}, and \texttt{HillClimb-Binary}. This still does not amount to a competition-grade baseline suite, but it is stronger and more domain-specific than the earlier ``demo-level'' comparison layer. The goal is not to claim exhaustive coverage of the literature, but to test whether MMAO remains credible once the external comparison set is no longer limited to only the lightest reproducible baselines.

\subsection{Budget Fairness}
Fairness is defined operationally in terms of matched task grids, seed grids, and explicit per-method budget accounting rather than wall-clock comparisons. The protocol therefore imposes:
\begin{itemize}
    \item matched seed sets across compared methods on each task;
    \item matched budget units within each task class;
    \item identical stopping criteria for all methods within a given comparison group;
    \item explicit reporting of initialization range, population constraints, and evaluation accounting.
\end{itemize}

For continuous optimization, the completed matrix uses one fixed configuration family for MMAO, with 110 iterations in 10D and 140 iterations in 30D, while the baselines are executed under fixed deterministic budgets: \texttt{RandomSearch} uses $80D$ objective evaluations, \texttt{HillClimb} uses $60D$ perturbation evaluations after initialization, \texttt{DE-lite} uses 70 generations with population 14, \texttt{PSO-lite} uses 60 iterations with swarm size 18, and \texttt{ES-lite} uses 45 iterations with an $(8,32)$-style update. For discrete optimization, MMAO uses 80 iterations with one shared configuration family, while the TSPLIB baselines use constructive initialization plus bounded 2-opt improvement and the \texttt{IG-2opt} baseline adds iterated perturb-and-improve cycles. The auxiliary multiple-knapsack slice uses matched iteration-style budgets for MMAO and matched constructive or local-improvement budgets for the external baselines. These rules support \emph{within-task comparability and reproducibility}, but they should not be misread as a claim that every primitive operation is exactly normalized across algorithm families.

\subsection{Statistical Reporting}
Each task-method cell reports mean, standard deviation, median, and best value. Beyond per-instance tables, the study also reports:
\begin{itemize}
    \item average rank across tasks for MMAO against the external baseline set;
    \item approximate Mann--Whitney comparisons between MMAO and each external baseline;
    \item effect-size style summaries for key ablation comparisons;
    \item trajectory-level summaries of communal budget, recent success, role-state drift, and population size on representative tasks.
\end{itemize}

This aggregate layer is necessary because benchmark papers can otherwise overstate performance from visually selective examples or isolated best-case wins. Given the current completed artifact set, the emphasis is placed on task-averaged summaries, average ranks, representative mechanism traces, and reproducible post-processing rather than on a larger family of omnibus tests.

\begin{figure}[t]
\centering
\includegraphics[width=\columnwidth]{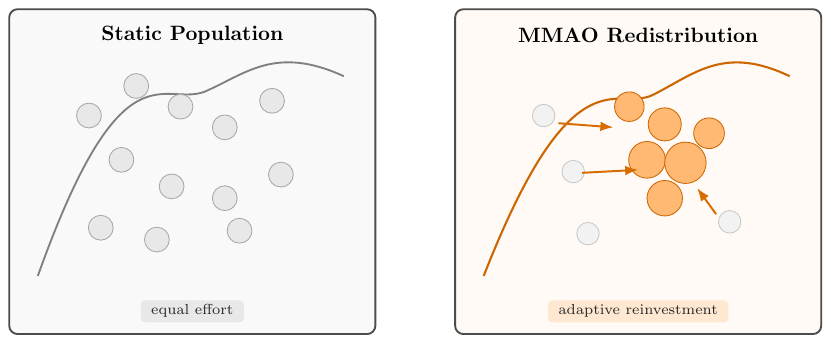}
\caption{Conceptual contrast between static equal-effort allocation and MMAO-style endogenous redistribution. The present study tests whether this principle remains beneficial under stricter benchmark pressure.}
\label{fig:comparison}
\end{figure}

\section{Experimental Results}
The benchmark campaign described in Section IV has been completed for the current evaluation matrix. We therefore report both task-level summaries and cross-task aggregate patterns. The discussion below is deliberately conservative: the present study evaluates a concrete, reproducible MMAO realization against stronger but still non-competition-grade baselines and against internal ablations. The conclusions apply to these baselines and ablations.

\subsection{Continuous Results}
Table~\ref{tab:continuous-baselines} summarizes the continuous comparison across all completed runs. The matrix contains eight CEC2017 functions at 10D and 30D, giving 16 task cells and 320 runs per method. MMAO substantially outperforms the external baseline set in aggregate. Its overall mean best-minus-bias is $5125.78$, with median $53.22$, compared with $2.61\times 10^8$ and median $592.12$ for PSO-lite, $7.22\times 10^8$ and median $1242.56$ for DE-lite, $1.51\times 10^9$ and median $3269.57$ for ES-lite, $5.78\times 10^9$ and median $14732.37$ for RandomSearch, and $7.00\times 10^9$ and median $18539.97$ for HillClimb. Table~\ref{tab:continuous-baseline-stats} makes the aggregate comparison more explicit through task-averaged ranks and approximate Mann--Whitney tests.

The dimension split is also informative. At 10D, MMAO attains mean best-minus-bias $441.44$ and median $33.62$, while at 30D the corresponding values rise to $9907.53$ and $215.99$. This confirms the expected pattern that the metabolic controller remains effective as dimensionality increases, but late-stage precision becomes more difficult to preserve at the larger scale. Even so, the degradation is far milder than for the stronger external baselines. For example, PSO-lite rises from mean residual $5.77\times10^7$ at 10D to $4.64\times10^8$ at 30D, while ES-lite rises from $5.72\times10^7$ to $2.96\times10^9$.

\begin{figure}[t]
\centering
\includegraphics[width=\columnwidth]{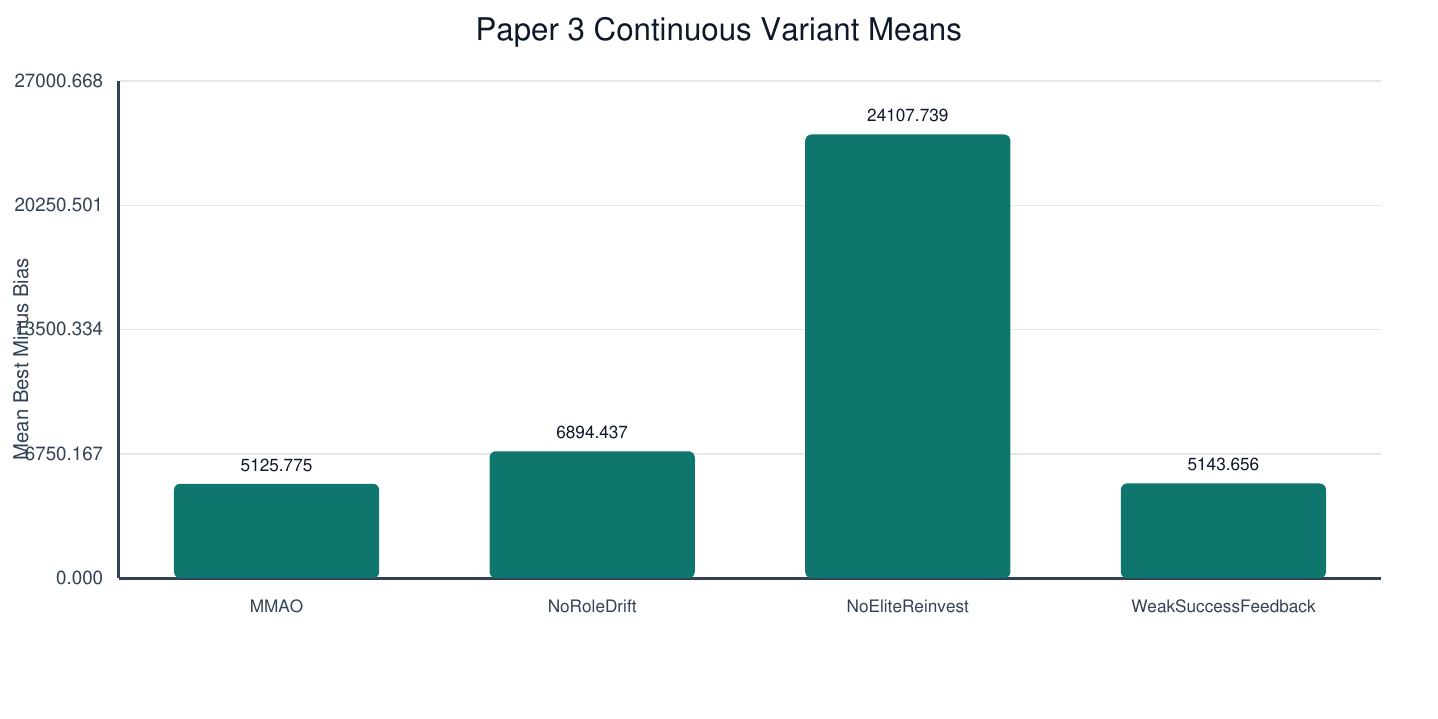}
\caption{Aggregate mean best-minus-bias for the continuous side. The current MMAO realization clearly improves over the external baseline set, while internal ablations remain relatively competitive.}
\label{fig:cont-convergence}
\end{figure}

At the task level, MMAO is particularly strong on \texttt{F52017} and \texttt{F92017}. The mean best-minus-bias on \texttt{F52017} is effectively zero, and \texttt{F92017} reaches negative values relative to the nominal bias. By contrast, larger residuals remain on \texttt{F102017} and \texttt{F12017}, indicating that hybrid and composition-like difficulty still exposes limitations in the present implementation. The aggregate cross-variant comparison is summarized in Fig.~\ref{fig:cont-convergence}.

\begin{table}[t]
\caption{Continuous benchmark summary. Lower is better.}
\label{tab:continuous-baselines}
\centering
\small
\resizebox{\columnwidth}{!}{%
\begin{tabular}{lcccc}
\toprule
Method & Runs & Mean & Std & Median \\
\midrule
MMAO & 320 & 5125.78 & 21311.05 & 53.22 \\
PSO-lite & 320 & $2.61\times 10^8$ & $1.13\times 10^9$ & 592.12 \\
DE-lite & 320 & $7.22\times 10^8$ & $2.87\times 10^9$ & 1242.56 \\
ES-lite & 320 & $1.51\times 10^9$ & $6.07\times 10^9$ & 3269.57 \\
RandomSearch & 320 & $5.78\times 10^9$ & $2.01\times 10^{10}$ & 14732.37 \\
HillClimb & 320 & $7.00\times 10^9$ & $2.57\times 10^{10}$ & 18539.97 \\
\bottomrule
\end{tabular}
}
\end{table}

\begin{table}[t]
\caption{Continuous aggregate baseline statistics. Lower rank and smaller objective residual are better.}
\label{tab:continuous-baseline-stats}
\centering
\small
\resizebox{\columnwidth}{!}{%
\begin{tabular}{lcccc}
\toprule
Method & Mean & Median & Avg. Rank & MW $p$ vs MMAO \\
\midrule
MMAO & 5125.78 & 53.22 & 1.063 & -- \\
PSO-lite & $2.61\times 10^8$ & 592.12 & 2.250 & $1.73\times 10^{-15}$ \\
DE-lite & $7.22\times 10^8$ & 1242.56 & 3.250 & $1.05\times 10^{-20}$ \\
ES-lite & $1.51\times 10^9$ & 3269.57 & 3.688 & $1.29\times 10^{-23}$ \\
RandomSearch & $5.78\times 10^9$ & 14732.37 & 5.188 & $1.36\times 10^{-40}$ \\
HillClimb & $7.00\times 10^9$ & 18539.97 & 5.563 & $7.80\times 10^{-47}$ \\
\bottomrule
\end{tabular}
}
\end{table}

\subsection{Discrete Results}
Table~\ref{tab:tsp-baselines} summarizes the TSPLIB side. The discrete matrix contains five instances and 20 seeds per instance, giving 100 runs per method. MMAO again clearly outperforms the external baseline set. Its overall mean gap is $7.84\%$, compared with $14.47\%$ for CI+2opt, $14.70\%$ for IG-2opt, $87.85\%$ for NN+2opt, and $405.01\%$ for RR-2opt. Table~\ref{tab:tsp-baseline-stats} complements the run-level summary with aggregate ranks and approximate Mann--Whitney tests.

\begin{figure}[t]
\centering
\includegraphics[width=\columnwidth]{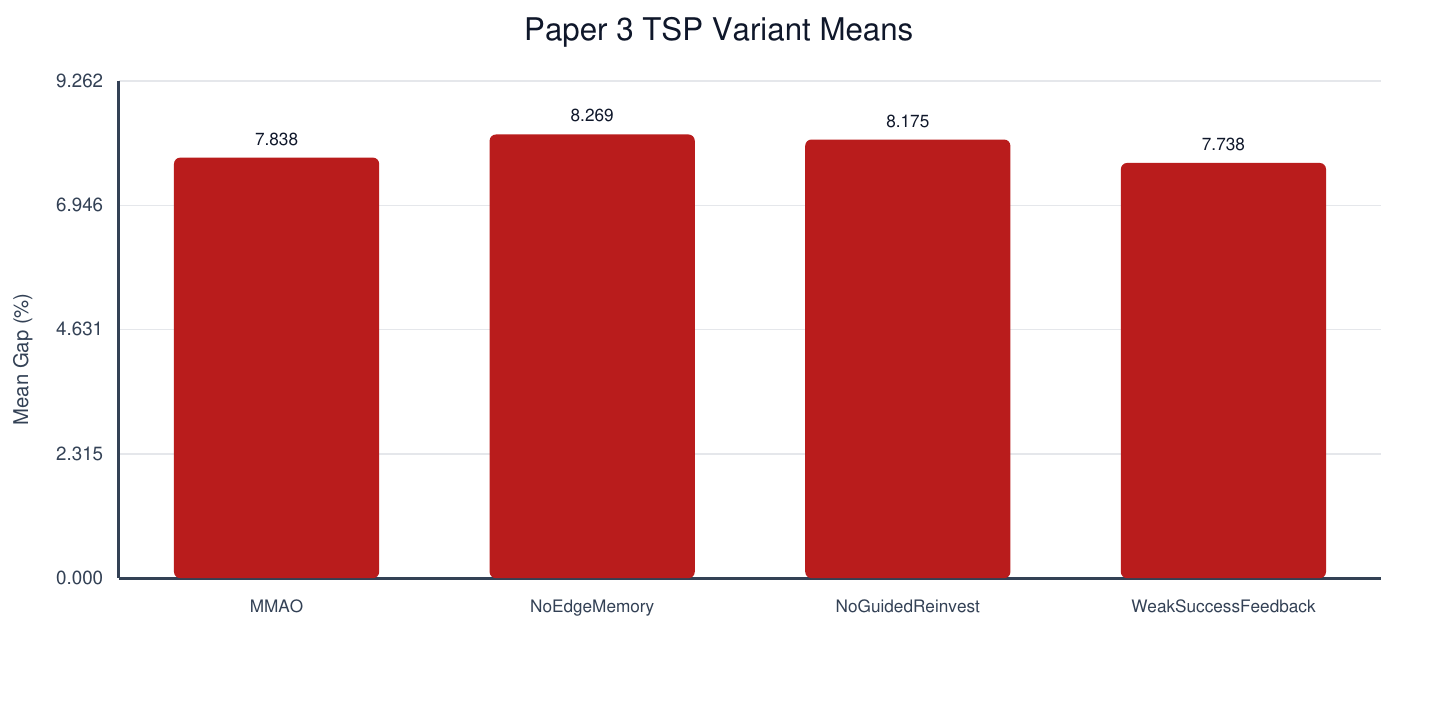}
\caption{Aggregate mean percentage gap for the TSPLIB side. MMAO clearly improves over the external TSP baselines, while the internal ablations remain close to the main variant.}
\label{fig:tsp-convergence}
\end{figure}

The task-level picture is more nuanced than the aggregate median alone suggests. MMAO is strongest on \texttt{eil51} and \texttt{berlin52}, where the mean gaps are $6.04\%$ and $6.35\%$, and remains reasonably stable on \texttt{st70}. Performance becomes looser on \texttt{eil76} and \texttt{kroA100}, where the mean gaps rise to $10.20\%$ and $8.65\%$. These values are still substantially better than the external TSP baselines, including \texttt{IG-2opt}, whose mean gaps are $11.79\%$, $18.41\%$, $12.79\%$, $12.89\%$, and $17.42\%$ on the five instances respectively. The aggregate discrete-side comparison is summarized in Fig.~\ref{fig:tsp-convergence}.

\begin{table}[t]
\caption{TSPLIB benchmark summary. Lower gap is better.}
\label{tab:tsp-baselines}
\centering
\small
\resizebox{\columnwidth}{!}{%
\begin{tabular}{lcccc}
\toprule
Method & Runs & Mean Gap (\%) & Std & Median \\
\midrule
MMAO & 100 & 7.84 & 2.92 & 7.42 \\
CI+2opt & 100 & 14.47 & 3.59 & 14.13 \\
IG-2opt & 100 & 14.70 & 3.60 & 14.68 \\
NN+2opt & 100 & 87.85 & 16.85 & 86.47 \\
RR-2opt & 100 & 405.01 & 151.18 & 362.36 \\
\bottomrule
\end{tabular}
}
\end{table}

\begin{table}[t]
\caption{TSPLIB aggregate baseline statistics. Lower rank and smaller percentage gap are better.}
\label{tab:tsp-baseline-stats}
\centering
\small
\resizebox{\columnwidth}{!}{%
\begin{tabular}{lcccc}
\toprule
Method & Mean Gap (\%) & Median Gap (\%) & Avg. Rank & MW $p$ vs MMAO \\
\midrule
MMAO & 7.84 & 7.42 & 1.000 & -- \\
CI+2opt & 14.47 & 14.13 & 2.400 & $3.28\times 10^{-25}$ \\
IG-2opt & 14.70 & 14.68 & 2.600 & $1.30\times 10^{-25}$ \\
NN+2opt & 87.85 & 86.47 & 4.000 & $2.52\times 10^{-34}$ \\
RR-2opt & 405.01 & 362.36 & 5.000 & $2.52\times 10^{-34}$ \\
\bottomrule
\end{tabular}
}
\end{table}

\subsection{Cross-Domain Reading}
A distinctive feature of the present study is that the continuous and discrete results can now be read under one reporting logic rather than as unrelated appendices. Two broad patterns emerge. First, MMAO clearly outperforms the external baseline set on both primary domains, which supports the claim that the metabolic resource loop produces useful adaptive behavior rather than merely metaphorical language. Second, the internal ablation outcomes are much closer to MMAO than the external baselines are. This is especially visible on the TSP side, where the full MMAO variant is clearly stronger than the external baselines but not decisively separated from every reduced variant. The most defensible interpretation is therefore framework-level credibility together with incomplete mechanism isolation.

The auxiliary OR-Library multiple-knapsack slice points in the same direction. On the currently completed slice, MMAO attains mean gap $0.085\%$ with median $0.014\%$, compared with $1.28\%$ for \texttt{GreedyDensity}, $23.81\%$ for \texttt{RandomRepair}, and $43.84\%$ for \texttt{HillClimb-Binary}. Its average rank on this auxiliary family is also $1.0$ against the external MKP baselines. Because this auxiliary family is smaller and not yet scaled to the same breadth as the TSPLIB campaign, we use it as supporting evidence rather than as the centerpiece of the discrete claim. Even so, it strengthens the statement that the closed-loop controller is not restricted to route construction alone.

The broader task matrices, stronger baselines, and repeated nonparametric summaries test whether endogenous redistribution remains useful under stricter evaluation. The results support the resource loop, while leaving the contribution of individual mechanisms partly unresolved.

\subsection{Direct Answers to RQ1--RQ5}
The completed matrix allows the five research questions from Section IV to be answered directly. For \textbf{RQ1}, the answer is positive within the present study scope: MMAO is consistently competitive on the selected CEC2017 subset and clearly outperforms the external baselines in aggregate, with average rank $1.063$ against the continuous baseline set and strongly favorable nonparametric comparisons. For \textbf{RQ2}, the answer is likewise positive on the discrete side: MMAO ranks first against the TSPLIB external baselines and also shows strong auxiliary behavior on the OR-Library multiple-knapsack slice.

For \textbf{RQ3}, the answer is more qualified. The 20-seed protocol shows that MMAO behavior is reproducible enough to maintain clear aggregate separation from the external baselines on both primary domains, but the per-task standard deviations also show that medium and difficult tasks still exhibit meaningful stochastic spread. For \textbf{RQ4}, the answer is partially positive rather than absolute: the ablation family supports the importance of the closed-loop framework, especially through the continuous-side reinvestment degradation, but it does not yet prove that every currently implemented submechanism is individually indispensable. For \textbf{RQ5}, the completed study reveals two main boundary conditions: late-stage precision on difficult higher-dimensional continuous tasks, and medium-scale discrete search where MMAO remains useful but not yet close to specialist-grade or exact-quality outcomes.

\section{Ablations and Mechanism Diagnostics}
To keep the strong-validation study aligned with the core MMAO claim, ablations must test closed-loop mechanisms rather than random implementation details. The present paper therefore implements a small but mechanism-centered ablation family on each side, and the results show that the full MMAO realization remains competitive, but its current advantage over nearby reduced variants is smaller than its advantage over the external baseline set.

\subsection{Primary Ablations}
The implemented ablations are:
\begin{itemize}
    \item \textbf{NoRoleDrift / Fixed role state:} removes continuous role drift;
    \item \textbf{NoEliteReinvest / NoGuidedReinvest:} weakens reinvestment on the continuous and discrete sides respectively;
    \item \textbf{WeakSuccessFeedback:} reduces the contribution of recent-success regulation;
    \item \textbf{NoEdgeMemory:} removes lightweight shared edge memory on the TSP side;
    \item \textbf{NoGuidedMemory:} removes shared memory guidance on the auxiliary multiple-knapsack side.
\end{itemize}

On the continuous side, the clearest degradation appears in \texttt{NoEliteReinvest}, whose overall mean best-minus-bias increases from $5125.78$ to $24107.74$, and whose 30D mean rises to $38616.34$. This is the strongest current evidence that reinvestment still matters materially in difficult continuous search. By contrast, \texttt{NoRoleDrift} and \texttt{WeakSuccessFeedback} remain close to MMAO in aggregate, and even achieve slightly better medians under the present summary statistics. On the discrete side, the differences are smaller still: the mean gaps of \texttt{NoEdgeMemory}, \texttt{NoGuidedReinvest}, and \texttt{WeakSuccessFeedback} remain close to the main MMAO variant on TSPLIB, while the auxiliary multiple-knapsack ablations also remain near the full method, with mean gaps of $0.097\%$ and $0.076\%$ respectively. Table~\ref{tab:ablation-stats} summarizes the corresponding aggregate ablation-side tests. The dominant pattern is that the present study separates MMAO clearly from external baselines, but only weakly from several nearby reductions. The implication is not that the ablated mechanisms are valueless, but that the current implementation and benchmark slice do not yet isolate them sharply enough.

\begin{table}[t]
\caption{Aggregate ablation statistics against MMAO. Smaller $p$ and larger $|\delta|$ indicate stronger separation.}
\label{tab:ablation-stats}
\centering
\small
\resizebox{\columnwidth}{!}{%
\begin{tabular}{llccc}
\toprule
Domain & Variant & Mean & MW $p$ vs MMAO & Cliff's $\delta$ \\
\midrule
Continuous & NoRoleDrift & 6894.44 & 0.9142 & 0.0049 \\
Continuous & NoEliteReinvest & 24107.74 & 0.0884 & -0.0778 \\
Continuous & WeakSuccessFeedback & 5143.66 & 0.9707 & -0.0017 \\
TSP & NoEdgeMemory & 8.27 & 0.3551 & -0.0757 \\
TSP & NoGuidedReinvest & 8.18 & 0.4688 & -0.0593 \\
TSP & WeakSuccessFeedback & 7.74 & 0.6895 & 0.0327 \\
MKP & NoGuidedMemory & 0.10 & 0.5390 & 0.0563 \\
MKP & WeakSuccessFeedback & 0.08 & 0.4265 & 0.0728 \\
\bottomrule
\end{tabular}
}
\end{table}

\subsection{Diagnostics}
The study includes representative trajectory-level diagnostics for communal budget, recent success rate, mean role state, and population size. These diagnostics are not intended as a full theory of MMAO dynamics, but they do let us check whether the closed-loop controller is behaving coherently rather than merely producing final scores.

\begin{figure}[t]
\centering
\includegraphics[width=\columnwidth]{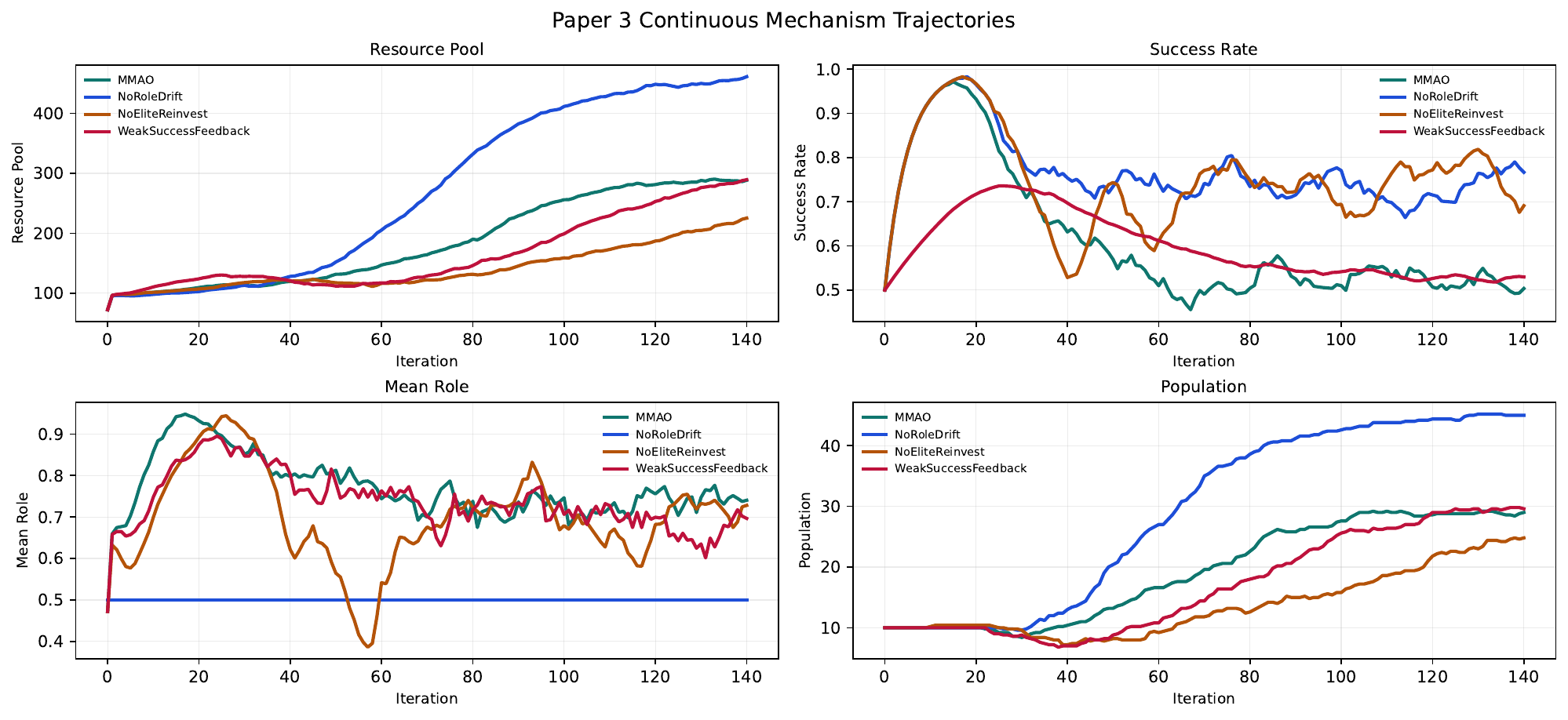}
\caption{Representative continuous-side mechanism trajectories. The communal pool, success rate, role state, and population size evolve differently across MMAO and selected ablations, which makes the metabolic controller more interpretable than a score-only comparison.}
\label{fig:cont-mech}
\end{figure}

Three patterns are especially visible. On the representative difficult continuous task in Fig.~\ref{fig:cont-mech}, the full MMAO variant finishes with elevated mean role state and a moderate communal pool, whereas \texttt{NoRoleDrift} freezes role movement and accumulates a much larger terminal pool together with a larger terminal population. Concretely, the terminal communal pool is about $301.8$ for MMAO versus $461.4$ for \texttt{NoRoleDrift}, while the mean role state ends near $0.74$ for MMAO and remains fixed at $0.5$ for \texttt{NoRoleDrift}; terminal population is about $30.4$ for MMAO versus $45.0$ for the ablated variant. This is consistent with the interpretation that role adaptation helps convert stored communal budget into directed effort rather than passive stockpiling. On the TSP representative task, the full MMAO variant ends with a slightly smaller population and substantially lower mean role than \texttt{NoGuidedReinvest}, while the recent success rate remains lower but stable ($0.342$ versus $0.421$). This suggests that the route-side controller tends to converge toward a leaner exploitation regime once useful structures are found, instead of preserving a broader active search front. On the auxiliary multiple-knapsack task, the terminal success rate falls sharply for all variants, with MMAO ending near $0.051$ and the two ablations near $0.076$ and $0.105$. That behavior is expected in a near-binary saturation regime, and the communal-pool, role, and population traces still remain well-behaved rather than exploding or collapsing numerically.

These trajectories describe controller behavior, with mechanism optimality remaining unresolved. The present study was designed first as an evidence paper rather than as a full internal-dynamics study. Completed score matrices, formal aggregate tables, nonparametric tests, ablation summaries, representative trajectory diagnostics, and reproducible post-processing already satisfy that evidential burden, while deeper causal mechanism analysis remains an important next step.

\section{Discussion and Threats to Validity}
The discussion must remain more cautious than the original MMAO introduction \cite{xu2026mmao}. A strong experimental study is valuable precisely because it clarifies both promise and boundary.

\subsection{Interpretation Scope}
The completed benchmark results support a stronger version of the original study hypothesis than was available at the draft stage: the closed-loop resource-allocation principle has clear empirical value across both continuous and discrete optimization when compared with the present external baseline set. That supports MMAO as a serious adaptive framework. At the same time, the ablation outcomes show that this evidence should be interpreted at the \emph{framework} level rather than as proof that every presently implemented submechanism is already mature or indispensable.

Benchmark breadth supports the cross-domain value of endogenous budget redistribution, while the ablations provide only partial evidence for the contribution of individual mechanisms. This distinction limits the conclusions that can be drawn from the current evaluation.

\subsection{Likely Failure Modes}
Several realistic limitations should remain visible even in a successful study:
\begin{itemize}
    \item continuous high-dimensional hybrid or composition functions still expose insufficient late-stage precision, especially through the large residuals on \texttt{F102017} and \texttt{F12017};
    \item TSPLIB evidence, while meaningful, does not establish credibility for all combinatorial optimization families, and the auxiliary multiple-knapsack slice is still smaller than the primary routing campaign;
    \item implementation quality still matters, so empirical weakness need not imply framework invalidity and empirical strength need not imply framework completeness;
    \item baseline coverage, though stronger than earlier MMAO studies and now broader than a purely lightweight comparison, is still lighter than a full-scale competition-grade comparison using L-SHADE, CMA-ES, ACO, tabu search, or LKH-class solvers.
\end{itemize}

Even with these boundaries, the study still performs an important methodological role. If a controller cannot remain competitive under this kind of fair-budget, repeated-trial pressure, then it is not a credible basis for later derivations or mechanism analysis. The present results clear that threshold, while also making clear where later domain-specific or theory-focused work still has to do additional lifting.

\subsection{Threats to Validity}
The most important threats are external validity, implementation fairness, and benchmark representativeness. The study controls these only partially. External validity is limited by the chosen eight-function CEC2017 subset, five-instance TSPLIB range, and the still modest auxiliary multiple-knapsack slice. Fairness is limited by differences in engineering maturity across baseline implementations, especially because the current baseline set is intentionally reproducible rather than competition-grade. Benchmark representativeness is therefore improved relative to a TSP-only discrete study, but it is still not broad enough to justify universal discrete claims. In addition, although the present artifact set now includes representative mechanism traces, those traces are still descriptive rather than causal and do not by themselves identify which local design choices are essential.

\section{Conclusion}
This paper serves as a dedicated validation study for MMAO under broader and stricter empirical conditions. We specified the evaluated MMAO configuration, executed a fair-budget benchmark matrix across continuous and discrete domains, defined stronger baseline families and mechanism ablations, and reported the results through repeated trials, aggregate tables, representative trajectory diagnostics, and nonparametric significance analysis.

The central claim remains intentionally modest. The completed results show that metabolically endogenous resource redistribution is a viable and interpretable cross-domain adaptive principle, and that the current MMAO realization clearly outperforms the implemented external baselines on both CEC-style continuous tasks and TSPLIB route-optimization tasks while also behaving strongly on an auxiliary OR-Library multiple-knapsack slice. At the same time, the ablation outcomes show that several reduced variants remain close to the full realization under aggregate summaries. The study provides benchmark evidence for MMAO across the tested domains, with further work needed on stronger competition-grade baselines, sharper mechanism isolation, and broader discrete validation.

The study provides an empirical reference for evaluating MMAO under stricter protocols, subject to the limitations in baseline coverage and mechanism isolation noted above.

\section*{Acknowledgement}
The related publications in Table~\ref{tab:mmao-research-series} provide complementary views of the MMAO research series.

\begin{table}[!ht]
\centering
\caption{Other publications in the MMAO research series.}
\label{tab:mmao-research-series}
\footnotesize
\setlength{\tabcolsep}{5pt}
\renewcommand{\arraystretch}{1.12}
\begin{tabular}{@{}p{\columnwidth}@{}}
\hline
\textbf{Related publication} \\
\hline
\emph{MMAO: A Metabolic Multi-Agent Optimizer with Endogenous Resource Allocation for Continuous and Discrete Optimization}~\cite{xu2026mmao}.\newline
\href{https://arxiv.org/abs/2606.28109}{arXiv:2606.28109}
\\[5pt]
\emph{Minimal MMAO: A Resource-Closed-Loop Framework for Adaptive Metaheuristic Search}~\cite{xu2026minimalmmao}.\newline
\href{https://arxiv.org/abs/2606.30450}{arXiv:2606.30450}
\\[5pt]
\emph{MMAO-Dyn: A Metabolic Multi-Agent Optimizer for Dynamic Optimization}~\cite{xu2026mmaodyn}.\newline
\href{https://arxiv.org/abs/2607.00846}{arXiv:2607.00846}
\\[5pt]
\emph{MMAO-Cls: Metabolic Multi-Agent Optimization for Joint Feature Selection and Classifier Tuning}~\cite{xu2026mmaocls}.\newline
\href{https://arxiv.org/abs/2607.01539}{arXiv:2607.01539}
\\[5pt]
\emph{Mechanism and Stability Analysis of Metabolic Closed-Loop Metaheuristics}~\cite{xu2026mechanism}.\newline
\href{https://arxiv.org/abs/2607.01551}{arXiv:2607.01551}
\\[5pt]
\emph{MMAO: A Unified Resource-Regulated Controller for Cross-Domain Optimization}~\cite{xu2026unifiedmmao}.\newline
\href{https://doi.org/10.1109/ACCESS.2026.3752083}{IEEE Access, 2026}
\\[5pt]
\hline
\end{tabular}
\end{table}

\bibliographystyle{IEEEtran}
\bibliography{Ref}

\begin{thebibliography}{10}
\providecommand{\url}[1]{#1}
\csname url@samestyle\endcsname
\providecommand{\newblock}{\relax}
\providecommand{\bibinfo}[2]{#2}
\providecommand{\BIBentrySTDinterwordspacing}{\spaceskip=0pt\relax}
\providecommand{\BIBentryALTinterwordstretchfactor}{4}
\providecommand{\BIBentryALTinterwordspacing}{\spaceskip=\fontdimen2\font plus
\BIBentryALTinterwordstretchfactor\fontdimen3\font minus
  \fontdimen4\font\relax}
\providecommand{\BIBforeignlanguage}[2]{{%
\expandafter\ifx\csname l@#1\endcsname\relax
\typeout{** WARNING: IEEEtran.bst: No hyphenation pattern has been}%
\typeout{** loaded for the language `#1'. Using the pattern for}%
\typeout{** the default language instead.}%
\else
\language=\csname l@#1\endcsname
\fi
#2}}
\providecommand{\BIBdecl}{\relax}
\BIBdecl

\bibitem{stripinis2024benchmarking}
L.~Stripinis, J.~K{\r{u}}dela, and R.~Paulavi{\v{c}}ius, ``Benchmarking
  derivative-free global optimization algorithms under limited dimensions and
  large evaluation budgets,'' \emph{IEEE Transactions on Evolutionary
  Computation}, vol.~29, no.~1, pp. 187--204, 2024.

\bibitem{vermetten2024large}
D.~Vermetten, C.~Doerr, H.~Wang, A.~V. Kononova, and T.~B{\"a}ck, ``Large-scale
  benchmarking of metaphor-based optimization heuristics,'' in
  \emph{proceedings of the genetic and evolutionary computation conference},
  2024, pp. 41--49.

\bibitem{cenikj2025comparing}
G.~Cenikj, G.~Petelin, and T.~Eftimov, ``Comparing optimization algorithms
  through the lens of search behavior analysis,'' in \emph{Proceedings of the
  Genetic and Evolutionary Computation Conference Companion}, 2025, pp.
  475--478.

\bibitem{xu2026mmao}
\BIBentryALTinterwordspacing
J.~Xu and L.~Ma, ``Mmao: A metabolic multi-agent optimizer with endogenous
  resource allocation for continuous and discrete optimization,'' 2026.
  [Online]. Available: \url{https://arxiv.org/abs/2606.28109}
\BIBentrySTDinterwordspacing

\bibitem{xu2026minimalmmao}
\BIBentryALTinterwordspacing
------, ``Minimal mmao: A resource-closed-loop framework for adaptive
  metaheuristic search,'' 2026. [Online]. Available:
  \url{https://arxiv.org/abs/2606.30450}
\BIBentrySTDinterwordspacing

\bibitem{holland1975adaptation}
J.~H. Holland, \emph{Adaptation in Natural and Artificial Systems}.\hskip 1em
  plus 0.5em minus 0.4em\relax Ann Arbor, MI: The University of Michigan Press,
  1975.

\bibitem{kirkpatrick1983optimization}
S.~Kirkpatrick, C.~D. Gelatt~Jr, and M.~P. Vecchi, ``Optimization by simulated
  annealing,'' \emph{science}, vol. 220, no. 4598, pp. 671--680, 1983.

\bibitem{li2020slime}
S.~Li, H.~Chen, M.~Wang, A.~A. Heidari, and S.~Mirjalili, ``Slime mould
  algorithm: A new method for stochastic optimization,'' \emph{Future
  generation computer systems}, vol. 111, pp. 300--323, 2020.

\bibitem{kiran2015tsa}
M.~S. Kiran, ``Tsa: Tree-seed algorithm for continuous optimization,''
  \emph{Expert Systems with Applications}, vol.~42, no.~19, pp. 6686--6698,
  2015.

\bibitem{eiben1999parameter}
A.~E. Eiben, R.~Hinterding, and Z.~Michalewicz, ``Parameter control in
  evolutionary algorithms,'' \emph{IEEE Transactions on Evolutionary
  Computation}, vol.~3, no.~2, pp. 124--141, 1999.

\bibitem{karafotias2015parameter}
G.~Karafotias, M.~Hoogendoorn, and A.~E. Eiben, ``Parameter control in
  evolutionary algorithms: Trends and challenges,'' \emph{IEEE Transactions on
  Evolutionary Computation}, vol.~19, no.~2, pp. 167--187, 2015.

\bibitem{kennedy1995particle}
J.~Kennedy and R.~Eberhart, ``Particle swarm optimization,'' in
  \emph{Proceedings of ICNN'95-international conference on neural networks},
  vol.~4.\hskip 1em plus 0.5em minus 0.4em\relax ieee, 1995, pp. 1942--1948.

\bibitem{Hansen2001}
N.~Hansen and A.~Ostermeier, ``Completely derandomized self-adaptation in
  evolution strategies,'' \emph{Evolutionary Computation}, vol.~9, no.~2, pp.
  159--195, 2001.

\bibitem{Tanabe2014}
R.~Tanabe and A.~S. Fukunaga, ``Improving the search performance of shade using
  linear population size reduction,'' in \emph{2014 IEEE Congress on
  Evolutionary Computation (CEC)}.\hskip 1em plus 0.5em minus 0.4em\relax IEEE,
  2014, pp. 1658--1665.

\bibitem{Stanovov2018}
V.~Stanovov, S.~Akhmedova, and E.~Semenkin, ``L-shade algorithm with rank-based
  selective pressure for solving solving cec 2017 benchmark problems,''
  \emph{Swarm and Evolutionary Computation}, vol.~43, pp. 115--131, 2018.

\bibitem{brest2006self}
J.~Brest, S.~Greiner, B.~Boskovic, M.~Mernik, and V.~Zumer, ``Self-adapting
  control parameters in differential evolution: A comparative study on
  numerical benchmark problems,'' \emph{IEEE Transactions on Evolutionary
  Computation}, vol.~10, no.~6, pp. 646--657, 2006.

\bibitem{omeradzic2024self}
A.~Omeradzic and H.-G. Beyer, ``Self-adaptation of multi-recombinant evolution
  strategies on the highly multimodal rastrigin function,'' \emph{IEEE
  Transactions on Evolutionary Computation}, 2024.

\bibitem{nesterov2017random}
Y.~Nesterov and V.~Spokoiny, ``Random gradient-free minimization of convex
  functions,'' \emph{Foundations of Computational Mathematics}, vol.~17, no.~2,
  pp. 527--566, 2017.

\bibitem{raponi2023optimizing}
E.~Raponi, N.~C. Rakotonirina, J.~Rapin, C.~Doerr, and O.~Teytaud, ``Optimizing
  with low budgets: A comparison on the black-box optimization benchmarking
  suite and openai gym,'' \emph{IEEE Transactions on Evolutionary Computation},
  vol.~29, no.~1, pp. 91--101, 2023.

\bibitem{antipov2024already}
D.~Antipov, B.~Doerr, and A.~Ivanova, ``Already moderate population sizes
  provably yield strong robustness to noise,'' in \emph{Proceedings of the
  Genetic and Evolutionary Computation Conference}, 2024, pp. 1524--1532.

\bibitem{Helsgaun2017}
K.~Helsgaun, ``An extension of the lin-kernighan-helsgaun tsp solver for
  constrained problems,'' Roskilde University, Roskilde, Denmark, Tech. Rep.,
  2017, lKH-3 version.

\bibitem{Voudouris1999}
C.~Voudouris and E.~P.~K. Tsang, ``Guided local search,'' \emph{Evolutionary
  Computation}, vol.~7, no.~1, pp. 1--27, 1999.

\bibitem{ye2023deepaco}
H.~Ye, J.~Wang, Z.~Cao, H.~Liang, and Y.~Li, ``Deepaco: Neural-enhanced ant
  systems for combinatorial optimization,'' in \emph{Advances in Neural
  Information Processing Systems}, vol.~36, 2023.

\bibitem{elorza2024transforming}
A.~Elorza, X.~Benavides, J.~Ceberio, L.~Hernando, and J.~A. Lozano,
  ``Transforming combinatorial optimization problems in fourier space:
  Consequences and uses,'' \emph{IEEE Transactions on Evolutionary
  Computation}, vol.~29, no.~4, pp. 977--989, 2024.

\bibitem{doerr2023understanding}
B.~Doerr and Z.~Qu, ``From understanding the population dynamics of the nsga-ii
  to the first proven lower bounds,'' in \emph{Proceedings of the AAAI
  Conference on Artificial Intelligence}, vol.~37, 2023, pp. 12\,408--12\,416.

\bibitem{zheng2024approximation}
W.~Zheng and B.~Doerr, ``Approximation guarantees for the non-dominated sorting
  genetic algorithm ii (nsga-ii),'' \emph{IEEE Transactions on Evolutionary
  Computation}, 2024.

\bibitem{doerr2025speeding}
B.~Doerr, M.~S. Krejca, and S.~Wietheger, ``Speeding up the nsga-ii via dynamic
  population sizes,'' in \emph{Proceedings of the AAAI Conference on Artificial
  Intelligence}, 2025, pp. 26\,964--26\,972.

\bibitem{li2022distributed}
J.-Y. Li, K.-J. Du, Z.-H. Zhan, H.~Wang, and J.~Zhang, ``Distributed
  differential evolution with adaptive resource allocation,'' \emph{IEEE
  transactions on cybernetics}, vol.~53, no.~5, pp. 2791--2804, 2022.

\bibitem{liu2022cooperative}
X.-F. Liu, J.~Zhang, and J.~Wang, ``Cooperative particle swarm optimization
  with a bilevel resource allocation mechanism for large-scale dynamic
  optimization,'' \emph{IEEE Transactions on Cybernetics}, vol.~53, no.~2, pp.
  1000--1011, 2022.

\bibitem{cho2025configx}
H.~Guo, Z.~Ma, J.~Chen, Y.~Ma, Z.~Cao, X.~Zhang, and Y.-J. Gong, ``Configx:
  Modular configuration for evolutionary algorithms via multitask reinforcement
  learning,'' in \emph{Proceedings of the AAAI Conference on Artificial
  Intelligence}, vol.~39, 2025, pp. 26\,982--26\,990.

\bibitem{zhang2025laos}
Y.~Zhang and G.~Yi, ``Laos: Large language model-driven adaptive operator
  selection for evolutionary algorithms,'' in \emph{Proceedings of the Genetic
  and Evolutionary Computation Conference}, 2025, pp. 517--526.

\bibitem{dong2025effective}
Z.~Dong and X.~Wang, ``Effective computational resource allocation in
  evolutionary multi-objective multi-task optimization,'' in \emph{2025 IEEE
  Congress on Evolutionary Computation (CEC)}.\hskip 1em plus 0.5em minus
  0.4em\relax IEEE, 2025, pp. 1--7.

\bibitem{li2014adaptivebandit}
K.~Li, {\'A}.~Fialho, S.~Kwong, and Q.~Zhang, ``Adaptive operator selection
  with bandits for a multiobjective evolutionary algorithm based on
  decomposition,'' \emph{IEEE Transactions on Evolutionary Computation},
  vol.~18, no.~1, pp. 114--130, 2014.

\bibitem{jiang2023knowledge}
Y.~Jiang, Z.-H. Zhan, K.~C. Tan, and J.~Zhang, ``Knowledge learning for
  evolutionary computation,'' \emph{IEEE transactions on evolutionary
  computation}, vol.~29, no.~1, pp. 16--30, 2023.

\bibitem{liu2023transfer}
X.-F. Liu, Z.-H. Zhan, and J.~Zhang, ``Transfer-based particle swarm
  optimization for large-scale dynamic optimization with changing variable
  interactions,'' \emph{IEEE Transactions on Evolutionary Computation},
  vol.~28, no.~6, pp. 1633--1643, 2023.

\bibitem{signorelli2025perturbation}
F.~Signorelli and A.~Yaman, ``A perturbation and speciation-based algorithm for
  dynamic optimization uninformed of change,'' in \emph{Proceedings of the
  Genetic and Evolutionary Computation Conference}, 2025, pp. 773--781.

\bibitem{xu2026mmaodyn}
\BIBentryALTinterwordspacing
J.~Xu and L.~Ma, ``Mmao-dyn: A metabolic multi-agent optimizer for dynamic
  optimization,'' 2026. [Online]. Available:
  \url{https://arxiv.org/abs/2607.00846}
\BIBentrySTDinterwordspacing

\bibitem{xu2026mmaocls}
\BIBentryALTinterwordspacing
------, ``Mmao-cls: Metabolic multi-agent optimization for joint feature
  selection and classifier tuning,'' 2026. [Online]. Available:
  \url{https://arxiv.org/abs/2607.01539}
\BIBentrySTDinterwordspacing

\bibitem{xu2026mechanism}
\BIBentryALTinterwordspacing
------, ``Mechanism and stability analysis of metabolic closed-loop
  metaheuristics,'' 2026. [Online]. Available:
  \url{https://arxiv.org/abs/2607.01551}
\BIBentrySTDinterwordspacing

\bibitem{xu2026unifiedmmao}
J.~Xu, B.~Li, L.~Ma, R.~Zhu, F.~Nie, and J.~Li, ``{MMAO}: A unified
  resource-regulated controller for cross-domain optimization,'' \emph{IEEE
  Access}, pp. 1--12, 2026.

\end{thebibliography}

\end{document}